\documentclass{amia}
\usepackage{graphicx}
\usepackage[labelfont=bf]{caption}
\usepackage[superscript,nomove]{cite}
\usepackage{csquotes}
\usepackage{wrapfig}
\usepackage[ruled,vlined,linesnumbered]{algorithm2e}
\usepackage{enumitem}
\usepackage{multirow}
\usepackage{caption}
\usepackage{subcaption}
\usepackage{todonotes}

\begin{document}

\title{Semantic Search for Large Scale Clinical Ontologies}

\author{Duy-Hoa Ngo, Madonna Kemp, Donna Truran, Bevan Koopman, Alejandro Metke-Jimenez}

\institutes{
    The Australian E-Health Research Centre, CSIRO, Australia \\
}

\maketitle

\noindent{\bf Abstract}
\textit{
Finding concepts in large clinical ontologies can be challenging when queries use different vocabularies. A search algorithm that overcomes this problem is useful in applications such as concept normalisation and ontology matching, where concepts can be referred to in different ways, using different synonyms. In this paper, we present a deep learning based approach to build a semantic search system for large clinical ontologies. We propose a Triplet-BERT model and a method that generates training data directly from the ontologies. The model is evaluated using five real benchmark data sets and the results show that our approach achieves high results on both free text to concept and concept to concept searching tasks, and outperforms all baseline methods.
}

\section*{Introduction}
\label{sec:Introduction}

Data standardisation is an important and challenging goal in the field of medicine \cite{Benson2016}. One of the key elements required to achieve semantic interoperability between clinical systems is the availability of common ontologies that define the concepts in the domain. Currently, the most comprehensive clinical ontology available is SNOMED~CT, which contains more than 340,000 concepts and has been widely adopted in Electronic Health Record (EHR) systems worldwide. However, despite the availability and coverage of large ontologies such as SNOMED~CT, there are still many challenges with adoption and implementation. These revolve around two main issues: 1) much of the source data is free text and needs to be mapped to a standard ontology; 2) existing systems use their own code systems and it is not feasible to replace them with SNOMED~CT. 

The first problem can be addressed by using natural language processing (NLP), whereby free text is analysed to identify relevant concepts. This process is called \textit{information extraction} \cite{Jurafsky2009} and it has been studied extensively in the biomedical domain. The process is typically divided in two phases: identifying the spans of text that represent relevant concepts and mapping these spans to concepts in a chosen ontology (also referred to as \textit{concept normalisation}).

The second problem can be addressed by mapping between ontologies. For any ontology of considerable size these maps cannot be generated manually and require at least partial automation. This area of research is called \textit{ontology matching}\cite{Euzenat2013}. Some of the most important elements that inform the matching process are the labels, synonyms and descriptions of the concepts.

Tailored solutions exists for both the concept normalisation problem and the ontology matching problem. However, these tend to be very specific to the particular use case. Instead, in this paper we propose a generalised method for searching large ontologies such as SNOMED~CT, using short spans of text as input. This method can be used generically for both concept normalisation and ontology matching. A novel algorithm is presented, based on deep-learning-derived word representations, that is capable of finding good candidate concepts, even in the absence of common vocabulary. Input to the algorithm can be free text, when doing concept normalisation, or a concept from a source ontology, when doing ontology matching. 

We empirically evaluate our method in a number settings --- both concept normalisation and ontology matching. The results show that our method can effectively find relevant concepts, outperforming a number of comparison baselines. In addition, we show that our method is particularly suited to finding concepts where the input shares little or no common terms with the relevant concept. 

\section*{Related Work}
\label{sec:RelatedWorks}

The problem of searching large ontologies has been studied in the context of data entry. Sevenster et al. \cite{sevenster2012} proposed and evaluated an autocompletion algorithm for large medical ontologies and showed that a multi-prefix matching approached performs better than the baseline approach that only completes the entered string to the right. A modified version of this algorithm is implemented in Ontoserver \cite{metke2018}, a high-performance FHIR terminology server, and it is used by default to do value set expansions\footnote{In FHIR, this is the operation used to implement auto-complete style widgets for data capture.}. However, this algorithm doesn't perform well on other tasks where the input strings are not partial prefixes but rather full words or short sentences. Also, the algorithm uses standard string matching so it only works well when the queries use the same vocabulary as the ontology being searched.

Searching ontologies has also been studied in the area of information extraction, specifically in the concept normalisation step where a span of text that has been identified as being relevant is mapped to a concept in an ontology. Wang et. al. \cite{wang2018} wrote an extensive literature review on clinical information extraction. An example of a state of the art algorithm specifically designed for the concept normalisation step can be found in the work of Luo et al. \cite{luo2019}. 

Finally, although not specific to ontology search, there is also relevant related work in the area of information retrieval (IR). Recently, advanced neural network methods have been developed to learn semantic representations of words and overcome the vocabulary mismatch problem of traditional IR models. Popular models that follow this trend include \textit{Word2Vec} \cite{Mikolov2013}, \textit{GloVe} \cite{Jeffrey2014} and \textit{fastText} \cite{Piotr2016}. Their underlying idea is based on the distributional hypothesis in linguistics, i.e., words that are used and occur in the same contexts tend to purport similar meanings\footnote{https://en.wikipedia.org/wiki/Distributional\_semantics}. Those neural networks are trained with large data resources by unsupervised learning algorithms. Once the training is finalised, every word located in the model's dictionary will be encoded by a fixed length embedding vector. Then, those embedding vectors can be used as inputs to a combination function (e.g., average function) or another neural network model to derive an embedding vector of a query or a document. A limitation of these methods is that once the training is completed, the embedding vectors are static, which means that an embedding vector of a given word is always the same regardless of the context of use. Therefore, they may face issues with polysemy when a word might have a different meaning in a specific context.

Several contextualised word embedding methods based on deep long short-term memory (LSTM) architectures, such as \textit{CoVe} \cite{Bryan2017}, \textit{ELMO} \cite{Matthew2018} and \textit{FLAIR} \cite{Akbik2018}, have been proposed to improve the understanding of words and sentences. The main difference with the static word embedding methods is that the words' embedding vectors are dynamically generated according to which context they have been used, i.e., surrounding words in a given sequence. Recently, \textit{transformer}-based approaches like \textit{BERT} \cite{Devlin2019}, \textit{XLNet} \cite{Zhilin2019}, \textit{RoBERTa} \cite{Yinhan2019} have been proposed and achieved  state-of-the-art results over most NLP downstream tasks. A key idea in these methods is that the meaning of a word in a sequence is represented by how much attention of that word attracts the other words in the sequence. Once the training of those models completes, they output a list of embedding vectors for all tokenised words of a given sequence. Then, those embedding vectors can be combined in different strategies to derive a semantic embedding vector for an input sequence.

A special feature of a search engine designed to search large ontologies is that the to-be-searched documents are concept labels, which are usually short and, therefore, sentence embedding methods are highly relevant because they can be used to compute semantic relatedness between this type of label. Recently, many sentence embedding methods, for example \textit{Doc2Vec} \cite{Quoc2014},  \textit{Skip-Thought} \cite{Ryan2015}, \textit{InferSent} \cite{Conneau2017}, \textit{Universal Sentence Encoder} \cite{Daniel2018} and the \textit{Sentence-BERT} model \cite{Reimers2019} have been proposed and have achieved good results in various natural language understanding tasks such as sentiment analysis, text classification, question answering and semantic textual similarity. \textit{Doc2Vec} is an extension of \textit{Word2Vec} that is trained with large, unlabelled text data. \textit{Skip-Thought} is another extension of the Skip gram \textit{Word2Vec} model that tries to predict the surrounding sentences of a given sentence. \textit{Universal Sentence Encoder} trains a transformer network which augments unsupervised learning whereas \textit{InferSent} trains a Siamese BiLSTM network with a max-pooling layer on top. Similar to the \textit{InferSent} architecture, \textit{Sentence-BERT} replaces a BiLSTM network with a BERT network and outperforms the other state-of-the-art methods on common semantic textual similarity and transfer learning tasks. These models have been trained on natural language inference data sets \cite{Samuel2015, Adina2018}, in which, an input to a learning model is a pair of sentences and the output is an inferred relation between them.

The algorithm we propose in this paper is most similar to \textit{Sentence-BERT} \cite{Reimers2019}. Sentence-BERT uses a Siamese architecture, and classification and regression objective functions. Our approach instead, uses a Triplet network \cite{Hoffer2015}, which processes three inputs in parallel, and a triplet loss function, which is a learning to rank metric for the three inputs.

\section*{Method}
\label{sec:Method}

In this section we provide the details of our semantic search engine model for large scale clinical ontologies. First, we give an overview of the model and its components, and explain how to rank results for a given query. Then, we outline our training procedure, optimization objective in developing the model.

\begin{figure}[htbp]
	\centering
		\includegraphics[width=0.9\textwidth]{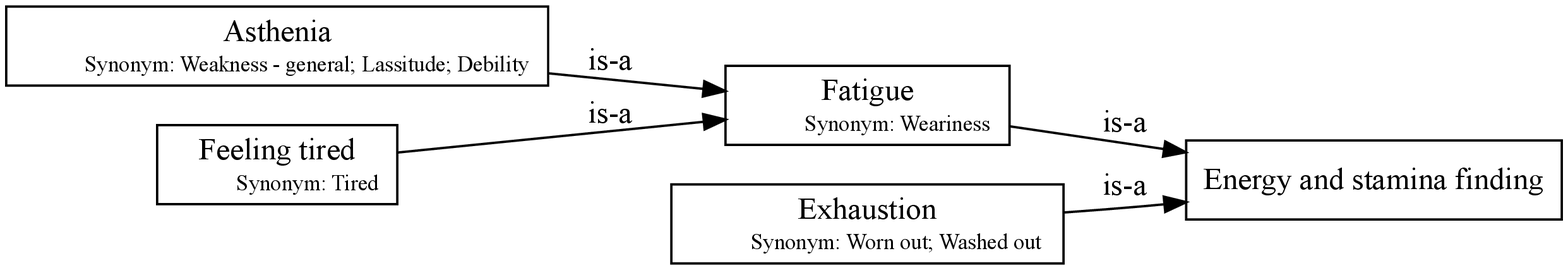}
	\vspace{-0.5em}
	\caption{A SNOMED CT fragment for Asthenia, Feeling tired, Fatigue, Exhaustion and Energy and stamina finding.}
	\label{fig:Fatigue}
\end{figure}

The main idea of this model is to transform every concept's label into appropriate embedding vectors in a vector space so their locations preserve the semantic relations between concepts in the ontology. Figure~\ref{fig:Fatigue} shows an example that illustrates this idea. The example shows that the concept \texttt{Asthenia} has three synonyms: \textquote{\textit{Weakness - general}}, \textquote{\textit{Lassitude}} and \textquote{\textit{Debility}}. Due to the characteristics of synonymy, we would expect that the distance between the embedding vectors of synonyms, e.g., \textquote{\textit{Asthenia}} vs. \textquote{\textit{Weakness - general}}, would be smaller than the distance between the vectors of labels of concepts that are not synonyms, e.g., \textquote{\textit{Asthenia}}vs. \textquote{\textit{Fatigue}} or \textquote{\textit{Exhaustion}}.

On the other hand, the concept \texttt{Feeling tired} is a sibling of the concept \texttt{Asthenia} because they both are children of the concept \texttt{Fatigue}. By applying the distance calculation method on the tree structure, the distance between two concepts is computed by the sum of the distances from those concepts to their lowest common ancestor. Therefore, we would also expect the  distance between embedding vectors of a concept's label to its direct parent concept's label to be smaller than the distance of that concept's label to its sibling concept's label. In this example, the distance between concept \texttt{Asthenia} and concept \texttt{Fatigue} must be smaller than the distance between concept \texttt{Asthenia} and its sibling \texttt{Feeling tired}. Again, because \textquote{\textit{Lassitude}} is a synonym label of \texttt{Asthenia} and \textquote{\textit{Weariness}} is a synonym label of \texttt{Fatigue}, we would infer that a embedding vector of \textquote{\textit{Lassitude}} is located closer to a embedding vector of \textquote{\textit{Weariness}} than to a embedding vector of  \textquote{\textit{Feeling tired}}. The intuition of distance comparison based on synonymy and tree-based distance can be applied to all concepts in an ontology.

\subsection*{Label embedding with Triplet-BERT model}
\label{sec:TripletBERT}

\begin{figure}[htbp]
	\centering
		\includegraphics[width=0.8\textwidth]{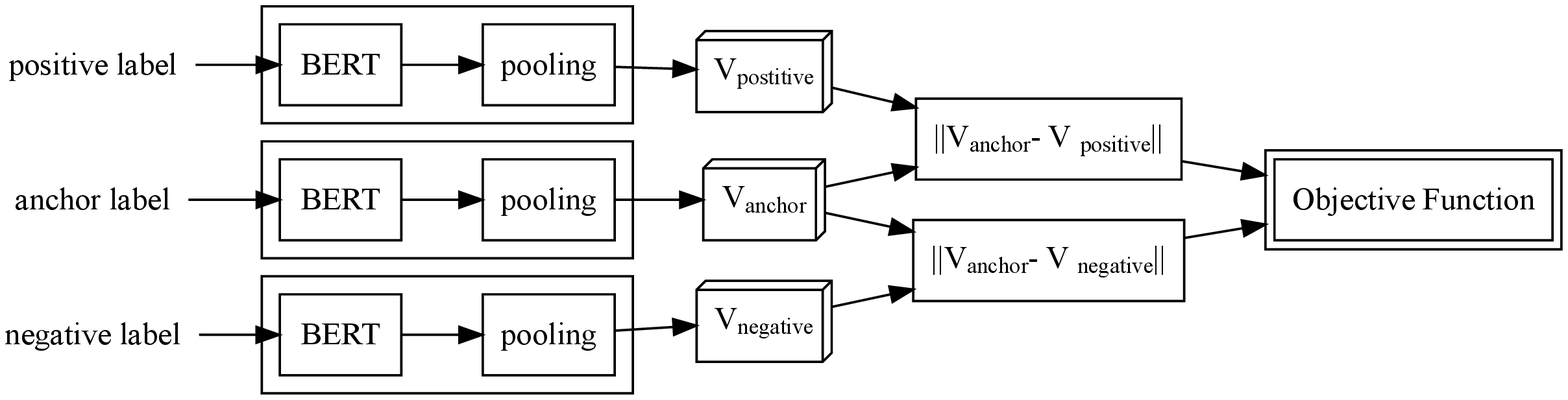}
	\vspace{-0.5em}
	\caption{Triplet-BERT model architecture}
	\label{fig:Triplet}
\end{figure}

In order to achieve a vector space model that observes the properties described in the previous section, a Triplet-Bert model was trained to produce a semantic embedding vector for short text spans such as concepts' labels or user queries. The Triplet-Bert model is a kind of Triplet network \cite{Hoffer2015} and it was mentioned by Reimers et al \cite{Reimers2019}. It consists of three instances of the same \textbf{embedding layer} containing a shared BERT network and a pooling layer (see Figure \ref{fig:Triplet}). It requires three text inputs, which are fed into the network at the same time, and represent three different roles: an \textbf{anchor} input, a \textbf{positive} input and a \textbf{negative} input. In this work, an \textbf{anchor} input is a user query (e.g., \textquote{\textit{Weakness - general}}); a \textbf{positive} input is a label of a high relevant concept (e.g., \textquote{\textit{Asthenia}}) to the user's query; and a \textbf{negative} input is a label of a less relevant concept (e.g., \textquote{\textit{Exhaustion}}) to the query. 

The objective of a Triplet-BERT model is to train its parameters so at the end, the encoded embedding vector of the \textbf{anchor} input is closer to the embedding vector of the \textbf{positive} input than that to the embedding of the \textbf{negative} input. Firstly, each input is fed into a BERT network to produce a list of intermediate embedding vectors, then, a pooling layer combines these vectors to produce a summary embedding vector for a given input. Now, after receiving three embedding vectors $V_{anchor}$, $V_{positive}$ and $V_{negative}$ for \textbf{anchor}, \textbf{positive} and \textbf{negative} inputs respectively, the model computes distances between the embedding vector of the anchor input against the embedding vectors of positive and negative inputs. Finally, the Objective Function module compares the two distances and indicates whether the model needs to adjust its parameters through the back propagation algorithm.  
Once the training has completed, the \textbf{embedding layer} is used to produce embedding vectors for any user text query as well as concept labels in the ontology. In order to rank results for a given query, the cosine similarity metric is used to compute relevance scores between the user queries and the concepts' labels.

\subsection*{Training Data Set}
\label{sec:TrainingDatasets}
Deep neural network methods had been recently used to develop sentence embedding models. Due to the huge number of parameters in deep neural networks, these models require a significant amount of training data. Some common published data sets used for training sentence embedding are the Stanford Natural Language Inference (SLNI) data set \cite{Samuel2015}, the Multi-Genre NLI data set (MLNI) \cite{Adina2018} and the Semantic Textual Similarity (STS) data set \cite{Daniel2017}. Each entry in those data sets consists of a pair of sentences and a label that is either a relationship type or a semantic similarity score. These entries cannot be used in our Triplet network because it requires three input sentences at the same time. Therefore, we propose a method to generate a data set for training our Triplet network (see Algorithm \ref{Alg:dataset}) based on the aforementioned intuition about distances of concepts in an ontology's hierarchy. The whole data set was generated from SNOMED~CT and the Human Phenotype Ontology (HPO). Each entry in the data set consists of three strings following the same order: an \textbf{anchor} label, a \textbf{positive} label and a \textbf{negative} label.  In total, the generated data set contains nearly \textbf{4 millions} entries, which are then split into \textit{training}, \textit{development} and \textit{testing} data sets with ratio of $90\%$, $5\%$ and $5\%$ respectively.

\begin{algorithm}[htbp]
\SetAlgoLined
\KwIn{T: Ontology}
\KwOut{$D_{train}$, $D_{dev}$, $D_{test}$}
D $\gets$ $\emptyset$\\
\ForEach{concept $\in$ T}{
    $conceptLabels$ $\gets$ getLabels($concept$, T) \\
    $directParents$ $\gets$ getParents($concept$, T)\\
    $otherConcepts$ $\gets$ getSiblings($concept$, T) $\cup$ getSiblings($directParents$, T)\\
    \ForEach{($label_{1}$ $\ne$ $label_{2}$) $\in$ conceptLabels}{
        $parentLabel$ $\gets$ getRandomLabel($directParents$, T)\\
        $otherLabel$ $\gets$ getRandomLabel($otherConcepts$, T)\\
        addToDataset(D, anchor=$label_{1}$, positive=$label_{2}$, negative=$parentLabel$)\\
        addToDataset(D, anchor=$label_{1}$, positive=$label_{2}$, negative=$otherLabel$)\\
        addToDataset(D, anchor=$label_{1}$, positive=$parentLabel$, negative=$otherLabel$)\\
    }
}
$D_{train}$, $D_{dev}$, $D_{test}$ $\gets$ splitTrainDevTest(D)
\caption{Generate training data from ontology for Triplet network}
\label{Alg:dataset}
\end{algorithm}


\subsection*{Training Details}
\label{sec:TrainingDetails}

\textbf{BERT network}. Transfer learning was used to fine-tune BERT parameters. In this work, we adopted BioBert-Base v1.1 \cite{Lee2019} --- a state-of-the-art biomedical language representation model, which has been widely using biomedical natural language processing tasks. 

\textbf{Pooling layer}. The pooling layer is added on top of the BERT network to get an embedding vector for a given text input. Different strategies can be used to work with BERT's output embedding vectors, however, according to Reimers et al \cite{Reimers2019}, the \texttt{MEAN} strategy achieved a better result than the others. Therefore, we chose the \texttt{MEAN} strategy for the pooling layer to produce a fixed size, i.e., a 768-dimensional embedding vector for the given text input.

\textbf{Distance metric}. We use Euclidean distance to compute distances between the anchor embedding vectors ($V_{anchor}$), the positive embedding vectors ($V_{positive}$) and the negative embedding vectors ($V_{negative}$).

\textbf{Objective function}. The objective of our model is to move the anchor embedding vector ($V_{anchor}$) closer to the positive embedding vector ($V_{positive}$) and far away from the negative embedding vector ($V_{negative}$). Therefore, we minimize the following objective function to tune the model's parameters:
\[ loss = \max(||V_{anchor} - V_{positive}|| - ||V_{anchor} - V_{negative}|| + m, 0)\]
Here, a small margin value $m$ is used to push the distance $||V_{anchor} - V_{negative}||$ being at least $m$ higher than the distance $||V_{anchor} - V_{positive}||$. Otherwise, the $loss\_value$ is positive, thus its derivatives to model' parameters are not 0, so the back propagation algorithm will update the model's parameters. In the training phase, we fix $m = 0.1$.

\textbf{Training settings}. Our model was trained in 5 epochs. We set a batch-size of 32, Adam optimizer with learning rate 2e-5, and a linear learning rate warm-up over 10\% of the training data. The training process was done in 40 hours with one GPU with 32G RAM, using Python 3.6, Pytorch 1.6 and CUDA 10.1. The Triplet-BERT codes was derived from Sentence-BERT \cite{Reimers2019} codes by replacing siamese network by triplet network.

\section*{Evaluation}
\label{sec:Evaluation}

In this section, we firstly describe how to evaluate the performance of our semantic search system and related evaluation metrics. Next, we present data sets used in the evaluation and finally we analyze the experimental results.

As our evaluation measure we use Hits@K. For a given query, Hits@K is 1 if the relevant concept is found in the top $K$ results; otherwise it is 0. In our evaluation, we used Hits@1, Hits@5 and Hits@10. Additionally, in order to measure the usefulness of the list of returned results for a given query, Normalized Discounted Cumulative Gain (nDCG) and Mean Reciprocal Rank (MRR) were used. Their underlying assumption is that the higher the relevant results are ranked, the more gain the user receives. Therefore, the \textit{ideal ranking} would first return the result with the highest relevance level, then the next highest relevance level, etc. nDCG@K measures the performance of a search system based on the relevance order of the $K$ returned results against the ideal ranking order. In our evaluation, we used nDCG@1, nDCG@5 and nDCG@10.

\subsection*{Data Sets for Evaluation}
\label{sec:DatasetEvaluation}

For evaluation, the following five data sets were used:
\begin{itemize}[noitemsep,topsep=0pt,parsep=0pt,partopsep=0pt]
  \item \textbf{cadec2sct}: This data set contains 2036 unique short text extracted and annotated to SNOMED~CT clinical finding concepts from medical forum posts on patient reported Adverse Drug Events (ADEs)\cite{Sarnaz2015}.
  \item \textbf{note2sct}: This data set contains 4960 unique short text extracted and annotated to SNOMED~CT clinical finding concepts from real patients' narrative discharge summaries in a hospital in Queensland, Australia.
  \item \textbf{hpo2sct}: This data set contains 14,149 unique labels collected from 5978 phenotype concepts from the Human Phenotype Ontology (HPO), which have been mapped to concepts in SNOMED~CT.
  \item \textbf{fma2sct}: This data set contains 13,123 unique labels collected from 5702 concepts from the Foundational Model of Anatomy Ontology (FMA), which have been mapped to concepts in SNOMED~CT.
  \item \textbf{ncit2sct}: This data set contains 46,185 labels collected from 13,830 concepts from the National Cancer Institute's Thesaurus (NCIt) which have been mapped to concepts in SNOMED~CT.
\end{itemize}

Three of the data sets, i.e., \textbf{cadec2sct}, \textbf{note2sct} and \textbf{hpo2sct} were manually annotated by two clinical terminology experts, whereas, the other two, i.e., \textbf{fma2sct} and \textbf{ncit2sct} were taken from the Large BioMed Track\footnote{http://www.cs.ox.ac.uk/isg/projects/SEALS/oaei/} in the Ontology Alignment Evaluation Initiative 2020\footnote{http://oaei.ontologymatching.org/}.

The first two data sets aim to evaluate the text-to-concept searching functionality where a user runs free-text queries to search relevant SNOMED~CT concepts. In the \textbf{cadec2sct} data set, clinical symptoms or diseases were described by different members with or without medical background, so the texts were not restricted to follow any standard naming rules. The entered texts were just the observation or understanding of lay people reporting adverse drug events. In contrast, in the \textbf{note2sct}, clinical entities were entered by doctors in hospital. Those texts are in free-text form, but the vocabulary is expected to be technical in nature. 

The last three data sets were extracted from well-designed biomedical ontologies. Thanks to cross-ontology alignment, it is expected that the results of searching a concept's label from one ontology will return its mapping concepts from another ontology that is found in the alignment. Additionally, these data sets can also be used to evaluate SNOMED~CT concept-to-concept search when all labels of a given concept from a source ontology, such as HPO, FMA or NCIt, are used to search relevant SNOMED~CT concepts.

\subsection*{Models for Comparison}
\label{sec:ModelComparison}

As presented in section \textbf{Method}, the key operation in our work is how to encode a short text into an embedding vector. For evaluation and comparison purposes, the following baseline embedding methods have been implemented to build different semantic search systems for SNOMED~CT.

\begin{itemize}[noitemsep,topsep=0pt,parsep=0pt,partopsep=0pt]
  \item \textbf{Elasticsearch BM25}: BM25 defines a weight for each term as a product of some IDF-function and some TF-function and then summarises that term weight as the score for the whole document towards the given query. In this work, a document is a collection of labels of a SNOMED~CT concept.
  \item \textbf{Word2Vec-based Average}: This model defines an embedding vector of a given text as an average of embedding vectors, which map to text's tokens in a pre-trained biomedical word2vec resource\footnote{https://bio.nlplab.org/}.
  \item \textbf{BioBERT based CLS}: This model uses a pre-trained BioBERT\footnote{https://github.com/dmis-lab/biobert} to encode an input text into a sequence of corresponding embedding vectors. Then, it defines an embedding vector of a given text by embedding a vector of special tokens [CLS] according to the original idea from BERT \cite{Devlin2019}.
  \item \textbf{BioBERT based MEAN}: Similar to the previous model but this model computes an average of all output embedding vectors as a summarised embedding vector for a given input.
  \item \textbf{Triplet-BERT}: We use the fine-tuned BioBERT with MEAN strategy for the pooling layer to encode a given text into an embedding vector.
\end{itemize}

The first model is a keyword-based search engine, which does not count synonymous features in its scoring function. The second model uses a pre-trained word embedding, trained on a very large biomedical text collection. Each word in a pre-trained word2vec always has the same embedding vector regardless of the context, so this model can be considered as a context-free embedding model. The third and the fourth models are based on a pre-trained BERT network, in which a word may have different embedding vectors depending on the context of use. The two models can be considered as general contextualised embedding models. The last model fine-tunes BERT parameters over the triplet data set generated from SNOMED~CT, and therefore it can be considered as a domain-specific contextualised embedding model.

\subsection*{Evaluation on Concept Normalisation Task}
\label{sec:Text2Concept}

In this experiment, a user provides a short query string and asks the system to return relevant concepts from SNOMED~CT. The query is either a clinical mention in \textbf{cadec2 sct} and \textbf{note2sct} or a concept's label in HPO, FMA and NCIt. We split the five data sets above into two groups based on the different editions of SNOMED~CT used in the ground truth data. The first group consists of three data sets including \textbf{cadec2sct}, \textbf{note2sct} and \textbf{hpo2sct}, where the ground truth data uses clinical finding concepts from the Australian edition of SNOMED~CT. The second group consists of two data sets, \textbf{nci2sct} and \textbf{fma2sct}, where the ground truth was created from concepts in the international edition of SNOMED~CT.

\begin{table}[htbp]
\small
\centering
\caption{Hits@K evaluation of searching a query to SNOMED CT clinical finding concepts.}
\vspace{-0.5em}
\label{Tab:cadecqueryhpo2sct}
    \begin{tabular}{ |l|c|c|c|c|c|c|c|c|c|} 
        \hline
         Dataset & \multicolumn{3}{|c|}{cadec2sct}  & \multicolumn{3}{|c|}{note2sct} & \multicolumn{3}{|c|}{hpo2sct} \\
        \hline
         Size & \multicolumn{3}{|c|}{\#queries=2,036}  & \multicolumn{3}{|c|}{\#queries=4,960} & \multicolumn{3}{|c|}{\#queries=14,149}\\
        \hline
         K & K=1 & K=5 & K=10 & K=1 & K=5 & K=10 & K=1 & K=5 & K=10\\
        \hline
        \hline
        BM25 & 0.132 & 0.240 & 0.307 & 0.526 & 0.705 & 0.770 & 0.334 & 0.486 & 0.553\\
        Word2Vec & 0.191 & 0.359 & 0.418 & 0.605 & \underline{0.778} & 0.829 & 0.397 & 0.571 & 0.632\\
        BERT-CLS & 0.114 & 0.216 & 0.283 & 0.406 & 0.553 & \underline{0.607} & 0.296 & 0.438 & 0.488\\
        BERT-MEAN & 0.136 & 0.266 & 0.319 & 0.255 & 0.472 & \underline{0.558} & 0.359 & 0.525 & 0.582 \\
        Triplet-BERT & \textbf{0.385} & \textbf{0.603} & \textbf{0.654} & \textbf{0.755} & \textbf{0.878} & \textbf{0.904} & \textbf{0.608} & \textbf{0.797} & \textbf{0.844}\\
        \hline
    \end{tabular}
\end{table} 

\begin{table}[htbp]
\small
\centering
\caption{Hits@K evaluation of searching a query to all SNOMED CT concepts}
\vspace{-0.5em}
\label{Tab:fmanci2sct}
    \begin{tabular}{ |l|c|c|c|c|c|c|} 
        \hline
         Dataset & \multicolumn{3}{|c|}{fma2sct} & \multicolumn{3}{|c|}{nci2sct}\\
        \hline
         Size & \multicolumn{3}{|c|}{\#queries=13,123} & \multicolumn{3}{|c|}{\#queries=46,185}\\
        \hline
         K & K=1 & K=5 & K=10 & K=1 & K=5 & K=10 \\
        \hline
        \hline
        BM25 & 0.310 & 0.571 & 0.639 & 0.345 & 0.459 & 0.496\\
        Word2Vec & 0.183 & \underline{0.506} & 0.597 & 0.369 & 0.513 & 0.567\\
        BERT-CLS & 0.157 & 0.244 & 0.287 & 0.296 & 0.378 & 0.412\\
        BERT-MEAN & 0.226 & 0.448 & 0.528 & 0.351 & 0.478 & 0.521\\
        Triplet-BERT & \textbf{0.700} & \textbf{0.855} & \textbf{0.885} & \textbf{0.503} & \textbf{0.638} & \textbf{0.670}\\
        \hline
    \end{tabular}
    \vspace{-10pt}
\end{table} 

Table \ref{Tab:cadecqueryhpo2sct} and Table \ref{Tab:fmanci2sct} show the results computing Hits@K values in the two groups with K = 1, 5 and 10 for the five aforementioned methods. Our method \textbf{Triplet\_BERT} consistently outperforms all other methods on all metrics over all five data sets. We have also computed the statistical significance \textbf{p-value} for Triplet-BERT on a paired \textbf{t-test} against the remaining methods. In almost all cases with various datasets and ranking value \textbf{K}, the computed value $p \ll 0.05$, which indicates strong evidence of significant differences between the results obtained by Triplet-BERT and the results returned from BM25, Word2Vec, BERT-CLS and BERT-MEAN. In Table \ref{Tab:cadecqueryhpo2sct} and Table \ref{Tab:fmanci2sct}, we highlighted cases where $p > 0.05$ by underlining the corresponding cell. For example, for dataset \textbf{note2sct} and $K=5$, the \textbf{p-value} computed for Triplet-BERT and Word2Vec is $0.75$. 

On the other hand, Triplet-BERT achieved a high Hits@10 value $0.904$ for data set \textbf{note2sct}, which means for a clinical mention written in the narrative of a discharge summary, the chance is about 90\% that annotators can find a correct SNOMED~CT concept within the top ten results that are returned. Similarly, the Hits@10 values are high for the \textbf{hpo2sct} and \textbf{fma2sct} data sets. Good results on these data sets were expected as the text in \textbf{note2sct}, \textbf{hpo2sct} and \textbf{fma2sct} was written by domain experts. In particular, the \textbf{note2sct} and \textbf{hpo2sct} data sets mainly focus on clinical finding concepts; the \textbf{fma2sct} data set mainly focuses on anatomical structure concepts, and both types of concepts are comprehensively covered by SNOMED~CT. 

The performance of \textbf{Triplet\_BERT } on \textbf{cadec2sct} and \textbf{nci2sct} datasets is a bit lower than on the \textbf{note2sct}, \textbf{hpo2sct} and \textbf{fma2sct} data sets. This is likely because of the quality of the text in \textbf{cadec2sct}. Many references were written in casual language which leads to irrelevant concepts being retrieved. For example, \textquote{threw up} was annotated to \texttt{422400008 | Vomiting (disorder) |}, but our system returned \texttt{282667008 | Does throw (finding) |}. Additionally, the lack of context also causes ambiguity; for example, the text \textquote{damage to my muscles} was annotated to \texttt{129565002 | Myopathy (disorder) |}, which means a disorder of skeletal and/or smooth muscle, but our system returned \texttt{95847005 | Injury of muscle (disorder) |}, which is a child concept of the annotated concept. 
The second reason for lower performance was posited as lower quality of synonymous labels of NCIt concepts in the \textbf{nci2sct} data set. For example, an NCIt concept \texttt{C1212} has the following labels: \textquote{sirolimus}, \textquote{rapamycin},  \textquote{SILA 9268a}, \textquote{WY-090217}, \textquote{AY 22989}, \textquote{rapamune} and \textquote{rapa}, but only the first two labels were found by our search system. The other labels contain either numeric tokens or abbreviations that do not provide helpful information to find relevant concepts.

An interesting observation here is that without the fine-tuning of \textbf{Triple\_BERT}, the basic \textbf{BERT} models did not outperform \textbf{Elastic BM25} and \textbf{Word2Vec} (even though one may expect them too). An explanation for this phenomenon is that the Elastic BM25 and Word2Vec mainly focus on keywords similarity, in which all \textit{stop-words} had been removed from the text during indexing and searching. On the contrary, the original BERT model creates embedding vectors for all tokens of the text, including \textit{stop-words}. In the case of BERT, due to the short query text and concept labels ($\approx$1-2 tokens), BERT's \textit{self-attention} layers may not capture the context of the text's tokens. For example, all the terms \textquote{\textit{Headache}}, \textquote{\textit{head pain}}, \textquote{\textit{Cephalodynia}}, \textquote{\textit{Cephalalgia}} and \textquote{\textit{Cephalgia}} refer to the same meaning - \textquote{\textit{pain in head}}. Pre-trained BioBERT computes similarity scores between (headache, cephalodynia) = 0.69; between (headache, cephalgia) = 0.73; and between (cephalodynia, cephalgia) = 0.97. The big difference in similarity scores tell us that in pre-train BioBERT, there is not enough context to embed those terms in highly similar vectors. This weakness is solved in \textbf{Triplet-BERT} because the model was fine-tuned from the original BERT model to push embedding vectors of those terms to be close to each other.

\subsection*{Impact of Overlapping Text and Synonymy in Searching Performance}
\label{sec:Overlapping}

In this experiment, we investigated the performance with respect to how similar queries are to their relevant concepts. Queries that are very similar to a relevant concept label will be easy to match, while queries that share no common terms will be harder to match. First, we define an overlapping degree as the proportion of shared tokens between query and concept label. Assuming that a query $q$ contains a list of \textit{non stop-words}: $T_q = \{t_{q1}, ... t_{qN}\}$, and similarly, the concept $c$ corresponding to the query $q$ in the ground truth data sets contains a list of \textit{non stop-words}: $T_c = \{t_{c1}, ... t_{cM}\}$ of all its labels, then an overlapping degree of $q$ against $c$ is calculated as follows:  $overlapping(q, c) = \frac{|| T_q \cap T_c  ||}{||T_q||}$. 
We divide our evaluation into two: 1) clinical findings concepts, which are often clearly expressed and have many synonym labels (\textbf{cadec2sct}, \textbf{note2sct} and \textbf{hpo2sct}); 2) all concepts types but with fewer synonym labels (\textbf{fma2sct} and \textbf{nci2sct}). We do this split to understand the impact that the synonym labels have for the different methods. 

\begin{figure}[htbp]
  \begin{subfigure}[b]{0.45\textwidth}
    \includegraphics[width=\textwidth]{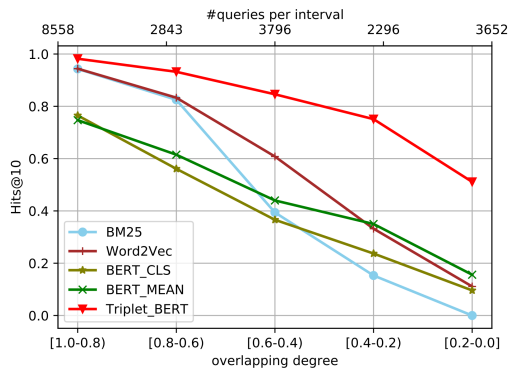}
    \caption{Evaluation on only clinical findings concepts with \textbf{more} synonym labels (cadec+note+hpo).}
    \label{fig:f1}
  \end{subfigure}
  \hfill
  \begin{subfigure}[b]{0.45\textwidth}
    \includegraphics[width=\textwidth]{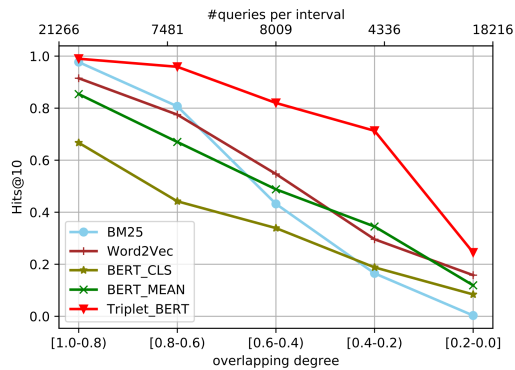}
    \caption{Evaluation on only clinical findings concepts with \textbf{fewer} synonym labels (fma+nci).}
    \label{fig:f2}
  \end{subfigure}
  \caption{Evaluation of searching performance with different overlapping degrees}
  \label{fig:Overlapping}
\end{figure}

Figure \ref{fig:Overlapping} shows the line charts of Hits@10 values of different methods at different intervals of overlapping degrees on the two aforementioned groups. (The left hand side of the plots shows queries with many shared terms; the right hand side shows queries with little or no term overlap.) As the overlapping degree goes down, the performance of all methods also decreases because these are harder queries to match. The line charts shows that the Elastic BM25 method achieved high performance when the queries share many terms with their corresponding concepts. Particularly, in the range of overlapping from $0.8-1.0$, the Hits@10 values of BM25 method was $0.943$ for the \textbf{cadec+note+hpo} group, and $0.977$ for the \textbf{fma+nci} group. Those values were close to the Hits@10 values of our Triplet-BERT method, i.e., $0.982$ and $0.990$ respectively. However, when the overlapping degree gradually decreases, the Hits@10 values of the BM25 method quickly falls to $0.0$. In contrast, Triplet-BERT still maintains its performance as the overlap decreases. The reason for this is that Triplet-BERT encodes the meaning of terms; thus it is still able to retrieve the relevant concept for a query even when it shares little or no common terms. Comparison between Figure~\ref{fig:f1} and Figure~\ref{fig:f2}, shows that number of synonym labels does not have a dramatic effect on performance for all five methods considered. A point to note, though, is that when Triple\_BERT has access to queries with more synonyms (Figure~\ref{fig:f1}), it maintains better performance for very low overlap queries compared with fewer synonyms (Figure~\ref{fig:f2}). This shows that Triple\_BERT does exploit synonyms for better performance.

Let us consider some specific examples of matching with low overlap queries. A query \textquote{narrow retinal arterioles} should match \texttt{271728000 | Retinal arteries attenuated (finding)|}. After running this query with five methods described above, we found that Elastic BM25 failed to return its correct result, whereas other methods were successful. An explanation is that the query and its corresponding concept shared only one word, \textquote{retinal}, which is $\frac{1}{3}$ of the query length. On the other hand, because \textquote{attenuated} vs. \textquote{narrow} as well as \textquote{arteries} vs. \textquote{arterioles} are semantically similar, the other methods, which rely on the meaning of the words, are able to find the correct result. Another interesting example is the query \textquote{tooth mass excess}, which matches SNOMED~CT concept \texttt{71485000 | Macrodontia (disorder)|}. The overlap degree is $0.0$. After running this query, only Triplet-BERT found the correct result.

\subsection*{Evaluation on the usefulness of the searching results}
\label{sec:EvaluationNDCG}

Search system are often evaluated according to the Normalized Discounted Cumulative Gain (nDCG) metric. We do the same here but first need to define the respective gain function a user receives for different types of results. Assume a user runs a query $q$. The `gain` $g$ a user receives for a result list $l$ against the correct result $q^{*}$ is defined as follows: $g=3$ if $l = q^{*}$; $g=2$ if $l$ is a direct parent or direct child of $q^{*}$ in the ontology; $g=1$ if $l$ is a grand parent, a grand child, a uncle or a sibling of $q^{*}$ and $g=0$ otherwise. 

\begin{table}[htbp]
\small
\centering
\caption{nDCG@K and Mean Reciprocal Rank (MRR) evaluation of searching methods}
\vspace{-0.5em}
\label{tab:nDCG}
    \begin{tabular}{ |l|c|c|c|c|c|c|c|c|} 
        \hline
         Datasets & \multicolumn{4}{|c|}{More synonyms (cadec+note+hpo)} & \multicolumn{4}{|c|}{Less synonyms (fma+nci)}\\
        \hline
         nDCG@K & K=1 & K=5 & K=10 & MRR & K=1 & K=5 & K=10 & MRR \\
        \hline
        \hline
        BM25 & 0.587 & 0.644 & 0.6646 & 0.425 & 0.540 & 0.585 & 0.586 & 0.401\\
        Word2Vec & 0.632 & 0.687 & 0.682 & 0.499 & 0.613 & 0.650 & 0.642 & 0.406\\
        BERT-CLS & 0.491 & 0.562 & 0.564 & 0.362 & 0.456 & 0.512 & 0.514 & 0.299\\
        BERT-MEAN & 0.520 & 0.596 & 0.595 & 0.372 & 0.540 & 0.589 & 0.586 & 0.385\\
        Triplet-BERT & \textbf{0.758} & \textbf{0.799} & \textbf{0.785} & \textbf{0.696} & \textbf{0.699} & \textbf{0.725} & \textbf{0.711} & \textbf{0.606}\\
        \hline
    \end{tabular}
\end{table} 

Table \ref{tab:nDCG} shows nDCG@1, nDCG@5, nDCG@10 and MRR values of different methods on two group datasets described in the previous section. By all metrics, Triplet-BERT method achieved the best performance. The average nDCG values produced by the Triplet-BERT is around 70\%, which means the order of the returned results is highly correlated to the order of ideal results.

Now, let's see an example to illustrate the importance of nDCG and MRR values. Assume that a user wants to find relevant SNOMED~CT concepts for the query \textquote{delayed closure of fontanels}. Our Triplet-BERT method returned a correct concept \texttt{82779003 | Late fontanel closure |} in the first rank;  the Word2Vec method returned this concept in second rank; both BERT-CLS and BERT-MEAN methods returned it in fifth position, whereas Elastic BM25 returned it in seventh position. That means that the user immediately finds the correct answer at the first or the second look if the system is based on Triplet-BERT or Word2Vec. In contrary, the user must spend more time to trace along the list of results to get a correct answer if the system uses BERT-CLS, BERT-MEAN or Elastic BM25. In this example, for Triplet-BERT: $MRR = \frac{1}{1}=1.0$; for Word2Vec: $MRR = \frac{1}{2}=0.5$; for BERT-CLS and BERT-MEAN: $MRR = \frac{1}{5}=0.2$ and for Elastic BM25: $MRR = \frac{1}{7}=0.14$. So, the higher value of $MRR$ is, the less time the user spends to find the correct answer for a given query.

Doing further analysis on this example, Triplet-BERT returns top five results including: the correct concept in the first position, its sibling concept \texttt{1667003} in the second position, its parent concept \texttt{248382004} in the third position, an uncle concept \texttt{249079005} in the fourth position and a grandparent concept \texttt{248381006} in the fifth position. Based on the respective gain defined above, Triplet-BERT obtains an $nDCG@5 = 0.976$. This number shows that the order of the returned results is very close to the ideal order, which means that the user can find not only the correct concept but also its close neighbours. 

Despite the fact that all five methods have found the correct results in their top 10 returned lists, which means the all have the same Hits@10 score, the order of returned results is different. This difference impacts the time that a user needs to find the correct answer. Our experiment shows that on average, the results obtained from Triplet-BERT were more useful than those from other baseline methods.

\subsection*{Evaluation on ontology matching task}
\label{sec:Concept2Concept}

In this experiment, we look for relevant concepts in an ontology for a given concept from  another ontology. This is ontology matching task, where each concept from an ontology can be mapped to one or several concepts from the other ontology. The concept-to-concept search slightly differs from the previous text-to-concept search in the way a the query is formulated. In concept normalisation (text-to-concept), a single query for a short text or a label was executed, whereas in concept-to-concept search, multiple queries, i.e., labels of a query concept, will be run. 

\begin{table}[htbp]
\small
\centering
\caption{Hits@K evaluation on concept to concept searching}
\vspace{-0.5em}
\label{tab:Concept2Concept}
    \begin{tabular}{ |l|c|c|c|c|c|c|c|c|c|} 
        \hline
         Dataset & \multicolumn{3}{|c|}{hpo2sct}  & \multicolumn{3}{|c|}{fma2sct} & \multicolumn{3}{|c|}{nci2sct}\\
        \hline
         Size & \multicolumn{3}{|c|}{\#concepts=5978}  & \multicolumn{3}{|c|}{\#concepts=5702} & \multicolumn{3}{|c|}{\#concepts=13830}\\
        \hline
         Hits@K & K=1 & K=5 & K=10 & K=1 & K=5 & K=10 & K=1 & K=5 & K=10\\
        \hline
        \hline
        BM25 & 0.529 & 0.718 & 0.778 & 0.454 & 0.776 & 0.831 & 0.598 & 0.755 & 0.793\\
        Word2Vec & 0.595 & 0.781 & 0.834 & 0.241 & 0.611 & 0.716 & 0.573 & 0.747 & 0.798\\
        BERT-CLS & 0.459 & 0.627 & 0.684 & 0.227 & 0.336 & 0.390 & 0.477 & 0.582 & 0.618\\
        BERT-MEAN & 0.549 & 0.740 & 0.794 & 0.302 & 0.576 & 0.680 & 0.541 & 0.693 & 0.740\\
        Triplet-BERT & \textbf{0.770} & \textbf{0.926} & \textbf{0.947} & \textbf{0.827} & \textbf{0.943} & \textbf{0.968} & \textbf{0.755} & \textbf{0.900} & \textbf{0.939}\\
        \hline
    \end{tabular}
\end{table} 

Table \ref{tab:Concept2Concept} shows the Hits@1, Hits@5 and Hits@10 values for concept normalisation. Triplet-BERT outperforms other methods on all metrics. If Hits@10 is considered, there is a $\approx 94\%$ chance that Triplet-BERT find a relevant SNOMED~CT concept be found for a given HPO, FMA and NCIt concept. A high recall at top 10 searching results can be used as an input to automate ontology matching~\cite{Hoa2016}. It would also greatly reduce the time and cost on manual ontology alignment.

\section*{Conclusion}
In this work, we proposed Triplet-BERT --- a label embedding model that can be used to build a semantic search system for large scale clinical ontologies. We also proposed a method for generating a training data set for the model directly from an ontology. The method is generic in nature and can be used for both concept normalisation and ontology matching. Several experiments were conducted using SNOMED~CT and showed that the proposed method outperforms baseline methods such as Elastic BM25, Word2Vec and BERT in all evaluation metrics on five benchmark data sets. In particular, the method was effective at mapping queries that had little or no common terms with relevant concepts. The strong empirical results suggest that Triplet-BERT can be used as the basis for both automatic ontology matching algorithms and searching tools to assist humans building ontology maps.

\makeatletter
\renewcommand{\@biblabel}[1]{\hfill #1.}
\makeatother

\bibliographystyle{unsrt}

\begin{thebibliography}{1}
\small
\setlength\itemsep{-0.1em}

\bibitem{Benson2016}
Benson, Tim and Grieve, Grahame.Principles of Health Interoperability SNOMED CT, HL7 and FHIR. Springer 2016.


\bibitem{Jurafsky2009}
Jurafsky, Daniel and Martin, James H.Speech and Language Processing (2nd Edition). Prentice-Hall 2009.

\bibitem{Euzenat2013}
Euzenat, Jrme and Shvaiko, Pavel. Ontology Matching. Springer 2013.

\bibitem{luo2019}
Yen-Fu Luo, et al. A Hybrid Normalization Method for Medical Concepts in Clinical Narrative using Semantic Matching. AMIA Joint Summits on Translational Science Proceedings 2019.

\bibitem{wang2018}
Yanshan Wang, et al. Clinical information extraction applications: a literature review. Journal of Biomedical Informatics 2018.

\bibitem{sevenster2012}
Merlijn Sevenster, Rob van Ommering and Yuechen Qian. Algorithmic and user study of an autocompletion algorithm on a large medical vocabulary. Journal of Biomedical Informatics 2012.

\bibitem{metke2018}
Alejandro Metke-Jimenez, Jim Steel, David Hansen, et al. Ontoserver: a syndicated terminology server. Journal of Biomedical Semantics 2018.

\bibitem{Reimers2019}
Reimers, Nils and Gurevych, Iryna. Sentence-BERT: Sentence Embeddings using Siamese BERT-Networks. EMNLP 2019.

\bibitem{Hoffer2015}
Elad Hoffer, Nir Ailon: Deep metric learning using Triplet network. ICLR 2015.

\bibitem{Ryan2015}
Ryan Kiros, Yukun Zhu, Ruslan R Salakhutdinov, Richard Zemel, Raquel Urtasun, Antonio Torralba, and Sanja Fidler. Skip-Thought Vectors. NeurIPS 2015.

\bibitem{Conneau2017}
Alexis Conneau, Douwe Kiela, Holger Schwenk, Loıc Barrault, and Antoine Bordes. Supervised Learning of Universal Sentence Representations from Natural Language Inference Data. EMNLP 2017.

\bibitem{Daniel2018}
Daniel Cer, Yinfei Yang, Sheng-yi Kong, et al. Universal Sentence Encoder. EMNLP 2018.

\bibitem{Lee2019}
Lee, Jinhyuk and Yoon, Wonjin et al. BioBERT: a pre-trained biomedical language representation model for biomedical text mining. Bioinformatics 2019.

\bibitem{Samuel2015}
Samuel R. Bowman, Gabor Angeli, Christopher Potts, and Christopher D. Manning. A large annotated corpus for learning natural language inference. EMNLP 2015.

\bibitem{Adina2018}
Adina Williams, Nikita Nangia, and Samuel Bowman. A Broad-Coverage Challenge Corpus for Sentence Understanding through Inference. NAACL 2018. 

\bibitem{Daniel2017}
Daniel Cer, Mona Diab, Eneko Agirre, Iigo Lopez-Gazpio, and Lucia Specia. Task 1: Semantic Textual Similarity Multilingual and Crosslingual Focused Evaluation. SemEval 2017.


\bibitem{Devlin2019}
Jacob Devlin, Ming-Wei Chang, Kenton Lee, Kristina Toutanova. BERT: Pre-training of Deep Bidirectional Transformers for Language Understanding. NAACL 2019.


\bibitem{Stephen2009}
Stephen Robertson, Hugo Zaragoza, et al. The probabilistic relevance framework: BM25 and beyond. TRIR 2009.

\bibitem{Mikolov2013}
Tomas Mikolov et al. Efficient Estimation of Word Representations in Vector Space. NeurIPS 2013.

\bibitem{Quoc2014}
Quoc V. Le and Tomas Mikolov. Distributed Representations of Sentences and Documents. ICML 2014.

\bibitem{Piotr2016}
Piotr Bojanowski et al. Enriching Word Vectors with Subword Information. TACL 2016.

\bibitem{Jeffrey2014}
Jeffrey Pennington. GloVe: Global Vectors for Word Representation. SIGDAT 2014.

\bibitem{Matthew2018}
Matthew E. Peters et al. Deep contextualized word representations. NAACL 2018.

\bibitem{Bryan2017}
Bryan McCann et al. Learned in Translation: Contextualized Word Vectors. NIPS 2017.

\bibitem{Akbik2018}
Alan Akbik et al. Contextual String Embeddings for Sequence Labeling. COLING 2018.

\bibitem{Yinhan2019}
Yinhan Liu et al. RoBERTa: A Robustly Optimized BERT Pretraining Approach. CoRR 2019.

\bibitem{Zhilin2019}
Zhilin Yang et al. XLNet: Generalized Autoregressive Pretraining for Language Understanding. NeurIPS 2019

\bibitem{Sarnaz2015}
Sarvnaz Karimi, Alejandro Metke-Jimenez et al. Cadec: A corpus of adverse drug event annotations. J Biomed Inform. 2015.

\bibitem{Hoa2016}
DuyHoa Ngo, Zohra Bellahsene. Overview of YAM++—(not) Yet Another Matcher for ontology alignment task. Journal of Web Semantics 2016. 


\end{thebibliography}

\end{document}